# Reinforcement Learning Based on Active Learning Method

Hesam Sagha[1], Saeed Bagheri Shouraki[2], Hosein Khasteh[1], Ali Akbar Kiaei[1]
[1]ACECR, Nasir Branch, Tehran, Iran
[2]Department of Electrical Engineering
Sharif University of Technology, Tehran, Iran
*sagha@ce.sharif.ir, bagheri-s@sharif.edu, h_khasteh@ce.sharif.ir, kiaei@ce.sharif.edu*

**Abstract**

*In this paper, a new reinforcement learning approach is proposed which is based on a powerful concept named Active Learning Method (ALM) in modeling. ALM expresses any multi-input-single-output system as a fuzzy combination of some single-input-single-output systems. The proposed method is an actor-critic system similar to Generalized Approximate Reasoning based Intelligent Control (GARIC) structure to adapt the ALM by delayed reinforcement signals. Our system uses Temporal Difference (TD) learning to model the behavior of useful actions of a control system. The goodness of an action is modeled on Reward-Penalty-Plane. IDS planes will be updated according to this plane. It is shown that the system can learn with a predefined fuzzy system or without it (through random actions).*

## 1. Introduction

ALM [1,2,3,4] is a recursive fuzzy algorithm, which expresses a multi-input-single-output system as a fuzzy combination of several single-input-single-output systems. It models the input-output relations for each input and combines these models to find out the overall system model. ALM starts with gathering data and projecting them on different data planes. The horizontal axis of each data plane is one of the inputs and the vertical axis is the output. IDS processing engine will look for a behavior curve, hereafter narrow line, on each data plane. If the spread of data over the narrow line is more than a threshold, data domains will be divided and the algorithm runs again. The heart of this learning algorithm is a fuzzy interpolation method which is used to derive a smooth curve among data points. It is done by applying a three-dimensional membership function to each data point, which expresses the belief for the data point and its neighbors. Each data point is considered as a source of light, which has a pyramid-shape illumination pattern. As the vertical distance from this source of light increases, its illuminating pattern will interfere with its neighbors forming new bright areas. The projection of the process on the plane is called IDS.

As it is shown in Fig 2, we can use a pyramid as a three dimensional fuzzy membership function of a data point and its neighboring points. By applying IDS method to each data plane, two different types of information will be extracted. One is the narrow path and the other is the deviation of the data points around each narrow path. Each narrow path shows the behavior of output relative to an input; and spread of the data points around this path shows the importance degree of that input in overall system behavior. Less deviation of data points around the path represents a higher degree of importance and vice versa.

Sagha *et al* [5] proposed a method which combines genetic algorithm and IDS to obtain better partitions over input variables. Their method is called GIDS (Genetic IDS).

Shahdi *et al* [6] proposed *RIDS* method that replaces each two consequent points with their midpoint instead of applying a 2-d fuzzy membership function on each data. *RIDS* converges to the center of gravity of data and increases the number of points in order to keep data expansion in plane. In addition, they proposed another method called (Modified *RIDS*) *MRIDS* that support negative data points. In *MRIDS,* if two consequent points are positive, the result is similar to that of *RIDS* and the replacing point is their midpoint. Nevertheless, if one of the points is negative, then the replacing point is a point located near the positive point on the line which connects two points; so negative point has an effect of deviating center of gravity from positive points.

*MRIDS* considers that the rewards and punishments are accessible after each action, but when they are delayed and this delay is not determined, it will not converge correctly.

Here we used another method called Reinforcement ALM (*RALM*), to add reinforcement capability to the algorithm. We used the concepts of Action Selection

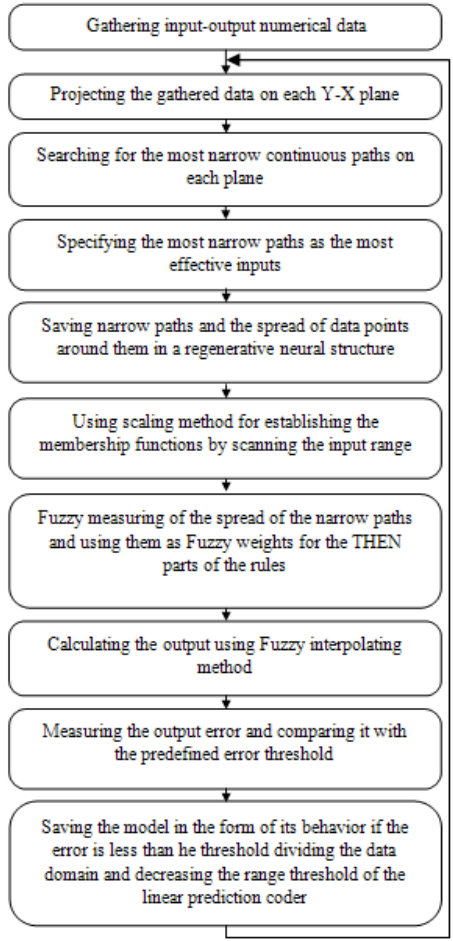

**Figure 1.** Flowchart of ALM

Network (*ASN*), Action Evaluation Network (*AEN*), and Stochastic Action Modifier (*SAM*) that are proposed in *GARIC* [7] as an actor-critic algorithm.

**GARIC:** The architecture of *GARIC* is schematically shown in Fig 3. *ASN* maps a state vector into a recommended action, *F*, using fuzzy inference. *AEN* maps a state vector and a failure signal into a scalar score that indicates state goodness. This is also used to produce internal reinforcement, *r′*. *AEN* can be a neural network structure or a fuzzy system [8] or alike. *SAM* uses both *F* and *r′* to produce an action *F′*, which will be applied to the plant. Learning occurs by fine-tuning the parameters in the two networks: in the *AEN*, the weights or fuzzy parameters are adjusted; in the *ASN*, the pa-rameters describing the fuzzy membership functions are changed. These are done by gradient descent approach. *AEN* parameters are updated via *Temporal Difference Learning* method.

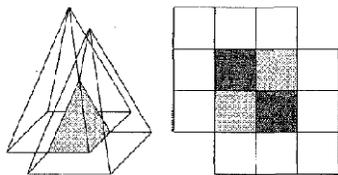

**Figure 2.** IDS method and Fuzzy membership function

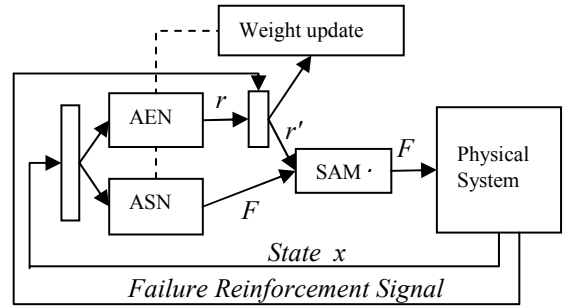

**Figure 3**. The artichecture of GARIC

**Temporal Difference Learning:** is a prediction method. It approximates its current estimate based on previously learned estimates by assuming subsequent predictions are often correlated in some sense. A prediction is made, and when the observarion is availabe the prediction is adjusted to better match the observation. If each state $s_t$ has the prediction value $v(s_t)$ that denotes the goodness of done actions in that state, then the updating formula is:

$$v(s_t) = v(s_t) + \alpha_t(R_{t+1} + \delta v(s_{t+1}) - v(s_t)) \quad (1)$$

where $\alpha_t$ is learning rate and $\delta$ is a constant in the range of [0,1] and $R_{t+1}$ is the received reward at time t+1.[9]

## 2. Proposed Method

In our method we used a similar structure to *GARIC*. *ASN* is an IDS fuzzy system. *AEN* is made up of a plane called Reward-Penalty-Plane (*RPP*). On this plane is the information of how much the done action in a specific state is good. From control viewpoint, this plane can be called *Error-Change in Error-Plane* because one axis denotes error and the other denotes changing error, and the value of each point in this plane shows that how much we can trust the selected action of *ASN* in that specific state. *SAM* changes the value of fuzzy system output by considering the output of *AEN*. More relevant an action is *SAM* changes it less. *RPP* is a surface made up of the desired variable to control. At first we have three regions on *RPP* plane, **i) reward area**: is the desired area we like the controller takes the state into that. This area has the fixed value of one**. ii) Penalty area**: consists of states that are not desirable in system and makes it unstable. The value of this area is stuck to the lowest negative value. **iii) Play area**: this area is the rest of the surface that has the value of zero initially. An initial plane is shown in figure 4, for a system we like to stable the variable *angle*.

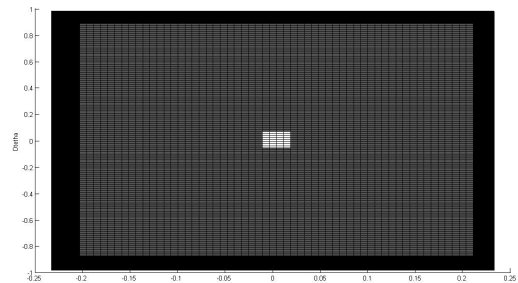

**Figure 4**. Initial Reward-Penalty Plane for an inverse pendulum system. Middle points denote the desired states (-0.012R< $\theta$ <0.012 R and -0.05R/s < $\Delta\theta$ < 0.05R/s) and have the maximum value (1), margin points denote penalty areas and have the negative minimum value (less than zero, more than -1), and other points are in the play area.

During the run when data is available, the value of *RPP* in the previous time step and its neighbors will approach to the value of *RPP* in the current time step and its neighbors as same as TD(0).

$$delta = \lambda win(RPP(e(t), ce(t)) - RPP(e(t-1), ce(t-1))) \quad (2)$$

$$RPP(e(t-1), e(t-1)) = RPP(ce(t-1), ce(t-1)) + delta \quad (3)$$

where, *delta* is the value of changing *RPP*, *e(t)* is the error of control variable at time *t*, *ce* is changing in error and *win* is the IDS window shows how much the neighbors must be effected and it can be a pyramid, Gaussian window with the center value of one or alike. When an action is rewarded the rate of update, $\lambda$, is high, but when an action is penalized, $\lambda$ is very low. It is because we assumed the number of false actions is much more than true actions for a system.

After updating the *RPP* plane, we must update IDS planes for the previous action and its neighbors. In this case, we reward the action if it goes to better state and punish it when it goes to worse state. The total goodness of an action will be obtained by averaging over delta values:

$$IDS_i(in_i(t-1), out(t-1)) = IDS_i(in_i(t-1), out(t-1)) + delta \quad (4)$$

where $in_i$ is the i[th] input variable.

Fuzzy system can be adapted online. In this case, after spreading each datum and neglecting data in negative areas of *IDS* planes, narrow lines of a predefined fuzzy inference system are updated. To select the next action (step time *t*) after fuzzy inference procedure in *ASN*, for exploration and exploitation in the space, we change the obtained value by following formula:

$$F(t) = ASN(t) + N(0, Var) \times (\exp(-\alpha \times RPP(e(t), ce(t))) - \exp(-\alpha)) \quad (5)$$

where *N(0,Var)* is a normally distributed random variable with variance *Var* and $\alpha$ is a constant. When *RPP* gives the best score for an action (i.e. 1), the selected action of *ASN* will be applied without manipulation.

For offline learning, after some data capturing, when the *RPP* plane converges and no changes occurred in it, or a specific number of iteration is passed, we get IDS planes that has both positive and negative values. Negative areas show that the actions which are chosen in this part of space transform the system into worse state. Therefore, by neglecting the data in these areas, we can filter bad actions. Finally, by estimating narrow lines, and using *ALM*, we can construct the fuzzy system.

Another problem exists when the state is in the range of reward area. If we use the original generated fuzzy system, we have vibrations in this range. It is because the learning system is not learned how to act when the state is in reward area. To handle this problem we used fuzzy scaling. In this kind of scaling, the range of input variable of fuzzy system will be scaled proportion to the reward range/input range:

$$In_i' = In_i \times Range(Reward(In_i)) / Range(In_i) \quad (6)$$

where $In_i$ is i[th] input which is a part of *RPP*.

Output range will be scaled by

$$Max(Range(Reward(In_i)) / Range(In_i)) \quad (7)$$

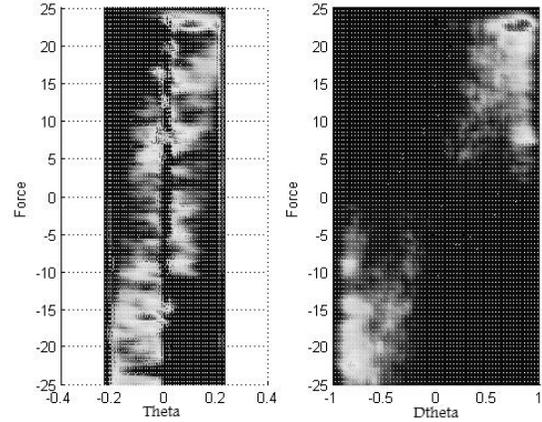

**Figure 5.** Final IDS planes for the inverted pendulum system. White areas have positive value and darker ones have negative values.

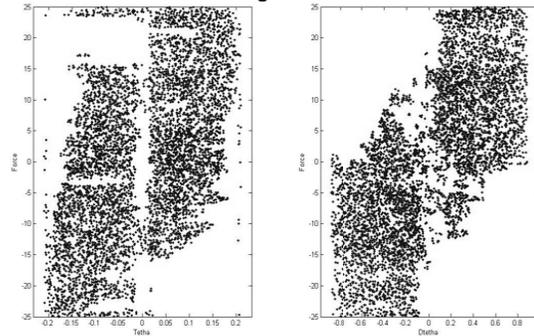

**Figure 6.** Selected data

Therefore, we do not need another fuzzy system; just variables must be scaled and use the generated fuzzy system.

This approach has some advantages in comparison with *MRIDS*, especially when the problem has delays to reach into a desired state. Also we explicitly define the reward and penalty areas and there is no need to define how and on what trajectory the system can reach the goal.

## 3. Results

We modeled the well-known inverse pendulum problem with two input, theta, $\theta$, and angular velocity (Dtheta), $\dot{\theta}$. Reward area was chosen between $\theta$ = [-0.23, 0.23] radian and $\dot{\theta}$ = [-0.98, 0.98] radian/s and penalty areas are when each variable is more than 0.9 of input range. $\lambda$ is chosen to be 0.9 for rewards and 0.05 for penalties. Penalty areas are set to be -0.5. Time step was chosen to be 0.021.

We used two methods of action selection in the beginning of run. The first one used random actions instead of a predefined fuzzy system, so no manipulation by formula (5) was needed. After 20000 sequences and only 32 successes during it, we got the IDS planes which are shown in Fig. 5. Success is when the system state is in the reward area. White areas are positive and dark areas are negative. Figure 6 shows the data that are extracted from all data after removing bad ones that are located on negative part of IDS planes. The final Reward-Penalty Plane is shown in Fig. 7. After applying ALM, we got a fuzzy system with only four rules. The surface of input-output-force is shown in Fig. 8. Fig. 9 shows some random initial states and their convergence to the desired point. Rise time is 2.71 and overshoot is %0.0. The second method uses an incorrect fuzzy system in ASN with four rules. After about 18000 time steps of online learning, the system learned to be stable. Fig 10 shows first 1000 time steps and last 1000 ones. Learning by GARIC takes about 100000 time steps but in our system it takes less than 20000 time steps.

In addition, we modeled ball and beam system. It is assumed the system has three inputs, $\theta$, $x0$, $v0$ and two outputs $x$ and $v$. $\theta$ is the angle of beam with horizontal line passing through the origin, $r0$ is the initial value of the distance of ball from the origin. $v0$ is the initial value of ball's speed and r and v are the final values of distance and speed. Our goal is to move the ball into the position zero, so we define the *RPP* with respect to $x$ and $v$. To control it, we have two inputs $v0$ and $x0$ and one output $\theta$.

The results of RALM for some random inputs are shown in Fig. 11. Generated fuzzy system has four rules. The rise time is 1.6 s and overshoot is %0.0. Table 1 shows the result of other proposed algorithm based on ALM and FALCON. It can be seen that rise time is reduced about 13% of supervised ALM and no overshoot is detected.

## 4. Conclusion

ALM is a powerful idea for modeling. We changed it to support reinforcement learning. Our approach uses another plane to get the information of reinforcement signals. The approach is useful when there is no explicit idea about the goodness of an action, and delayed rewards and penalties must be considered. *RALM* considers these very well.

Results show that RALM learns better than other proposed ALM based algorithms.

**Table 1.** Comparing control parameters of 4 controlling method

|  | Fuzzy rules | Over-shoot | Rise time |
|---|---|---|---|
| FALCON-ART | 28 | 23.5% | 2.11 |
| Unsupervised ALM | 4 | 14.3% | 1.87 |
| Supervised ALM | 4 | 0 | 1.85 |
| RALM | 4 | 0 | 1.61 |

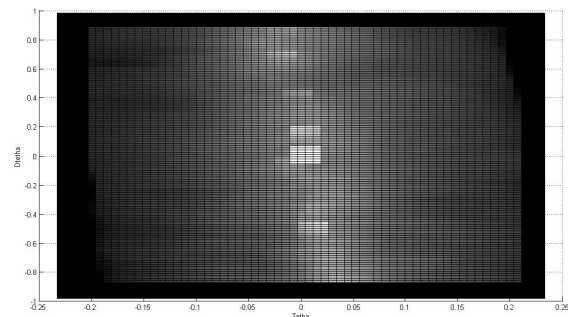

**Figure 7**. Reward--Plane

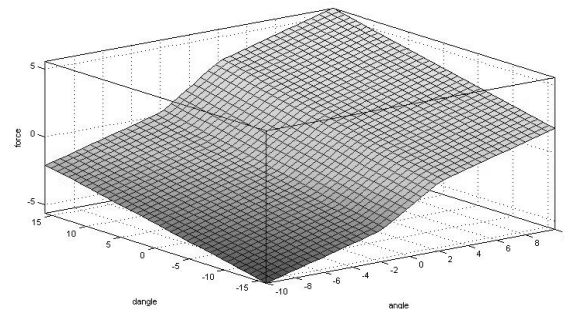

**Figure 8.** Fuzzy system surface

## 5. References

[1] S.Bagheri, G.Yuasa, N.Honda, "Fuzzy Controller Design by an active Learning Method",*31st Symposium of Intelligent Control,* SIC 98

[2] S.Bagheri, N.Honda,"Hardware Simulation of Brain Learning Process", *15th Fuzzy Symposium*, Osaka, June 99.

[3] S.Bagheri, N.Honda, "A New Method for Establishing and Saving Fuzzy Membership Functions", *13th Fuzzy symposium*, Toyama, 1997.

[4] S.Bagheri, N.Honda, "Outlines of a Soft Computer For Brain Simulation", *Methodologies for the Conception, Design And Application of Soft Computing, IIZUK*, 1998

[5] H.Sagha, S. B. Shouraki, M.Dehghani, "Genetic Ink Drop Spread", *International Symposium in Intelligent Information Technology Application, IITA,* China,2008

[6] A.Shahdi, S.Bagheri, "Supervised Active Learning Method as an intelligent linguistic Controller and Its Hardware Implementation", *IASTED*, Spain, 2002

[7] H.Berenji, P.Khedkar, "Learning and Tuning Fuzzy Logic Controllers Through Reinforcement", *IEEE Transaction on Neural Network*, Vol 3, No 5, 1992.

[8] H.Berenji, P.Khedkar, "Using Fuzzy Logic For Performance Evaluation in Reinforcement Learning", NASA-TM-111486

[9] R. Sutton,A. Barto. *Reinforcement Learning.* MIT Press, 1998

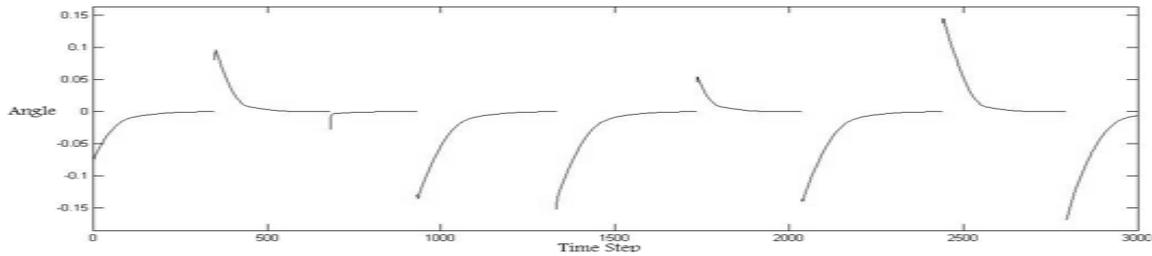

**Figure 9.** Final system output for inverse pendulum.

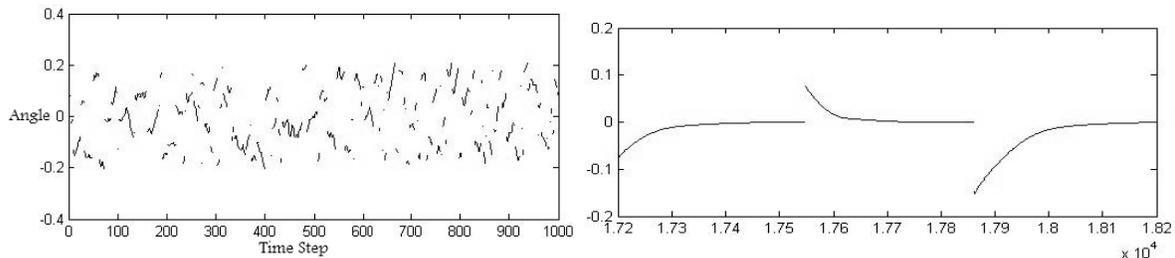

**Figure10.** Online learning; Left: First 1000 time steps, Right: Time steps between 17200 and 18200

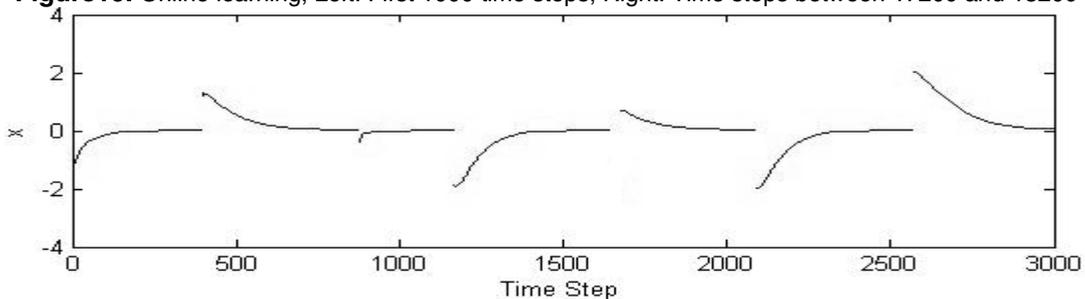

**Figure 11.** Final system output for ball and beam problem

602